\documentclass[submission,copyright,creativecommons]{eptcs}

\providecommand{\event}{The 9th International Symposium on Symbolic Computation in Software Science}

\usepackage{amsthm}
\usepackage{amsmath}
\usepackage{amssymb}
\usepackage{ifthen}
\usepackage{graphicx}
\usepackage{xcolor}
\usepackage{dsfont}
\usepackage{graphicx}
\usepackage[noend]{algorithmic}
\usepackage{algorithm}
\usepackage{underscore}
\usepackage{url}

\graphicspath{ {./images/} }

\newcommand{\newlengthsettowidth}[2]{\newlength {#1} \settowidth{#1}{#2}}

\newcommand{\numberRange}[1]{\mathbb #1}
\newcommand{\RR}{\numberRange R}                        
\newcommand{\BB}{\{0,1\}}                               

\newcommand{\range}[3][X]{\Ifthen{\Equal{#1}{X}}{\{}#2,\ldots,#3\Ifthen{\Equal{#1}{X}}{\}}} 

\newcommand{\atl}{\geq}                                 
\newcommand{\atm}{\leq}                                 

\newcommand{\argmax}{\operatorname{argmax}}

\newcommand{\func}[3]{#1 \colon #2 \rightarrow #3}      

\newcommand{\Ifthenelse}[3]{\ifthenelse{#1}{#2}{#3}}   
\newcommand{\Ifthen}    [2]{\Ifthenelse{#1}{#2}{}}

\newcommand{\Equal}     [2]{\equal{#1}{#2}}            

\newcommand{\True}         {\Equal{1}{1}}

\providecommand{\Draftmode}{\True} 

\newcommand{\Itedraft}    [2]{\Ifthenelse{\Draftmode}{#1}{#2}}
\newcommand{\Ifdraft}     [1]{\Itedraft{#1}{}}

\newcommand{\drafttext}   [2][\draftcolor]{\Ifdraft{{\color{#1}#2}}}
\newcommand{\draftpointer}{\makebox[0mm][c]{$^{*}$}}
\newcommand{\draftmargin} [2][\draftcolor]{\drafttext[#1]{\draftpointer\marginpar[\raggedleft\small{\color{#1}#2}]{\raggedright\small{\color{#1}#2}}}} 

\newcommand{\draftappendix}[1][Draft Appendix]{\Itedraft{\clearpage \section*{#1}}{\end{document}}}

\newcommand{\draftauthor}[3]{
  \newcommand{#1}[1]{\drafttext  [#3]{##1}}  
  \newcommand{#2}[1]{\draftmargin[#3]{##1}}  
}

\newcommand{\Ifspacecolor}{gray}
\draftauthor{\Ifspace}{\Ifspacemargin}{\Ifspacecolor}

\Ifdraft{\setlength{\overfullrule}{10mm}}

\newcommand{\centerTwoOut}  [2]               {#1 \hfill #2}                          

\newlength{\posLength}

\newcommand{\mc}{\multicolumn}                          
\newlengthsettowidth{\tabLength}{\ \ \ }

\newcommand{\wbox}[2][\tabLength]{\hspace*{#1}\mbox{#2}\hspace*{#1}}

\newcommand{\End}{\end{document}}                       
\newcommand{\emphasize}[1]{\textbf{#1}}                 

\newcommand{\refCapitalOrSmall}[3]{#1#3} 

\newcommand{\refFullOrAbbrev}[2]{#1} 

\newcommand{\algorithmref}  [2][!]{\genericref[#1] A a {\refFullOrAbbrev{lgorithm}  {lg.}}  {algorithm}  {#2}}

\newcommand{\figureref}     [2][!]{\genericref[#1] F f {\refFullOrAbbrev{igure}     {ig.}}  {figure}     {#2}}

\newcommand{\lineref}       [2][!]{\genericref[#1] L l {\refFullOrAbbrev{ine}       {ine}}  {line}       {#2}}

\newcommand{\sectionref}    [2][!]{\genericref[#1] S s {\refFullOrAbbrev{ection}    {ec.}}  {section}    {#2}}
\newcommand{\tableref}      [2][!]{\genericref[#1] T t {\refFullOrAbbrev{able}      {able}} {table}      {#2}}

\newcommand{\equationref}[2][!]{\Ifthen{\Equal{#1}!}{\refCapitalOrSmall E e {\refFullOrAbbrev{quation}{q.}}~}(\ref{equation: #2})}

\newcommand{\genericref} [6][!]{\Ifthen{\Equal{#1}!}{\refCapitalOrSmall{#2}{#3}{#4}~}\ref{#5: #6}} 
 \renewcommand{\refFullOrAbbrev}[2]{#2} 

\newcommand{\defFullOrAbbrev}[2]{#1} 

\newlength{\spaceAfterEOP}
\newcommand{\Proof}{\noindent\textbf{Proof}}               

\algsetup{indent=1.3em}

\newcommand{\plkeyword}[1]{\textbf{#1}}

\newcommand{\plreturn} {\plkeyword{return}}

\newcommand{\code}[1]{\texttt{#1}}

\newcommand{\data} {\mathbf{X}} 
\newcommand{\weights}{W} 
\newcommand{\poi}{w} 
\newcommand{\sample}{X} 
\newcommand{\lcomb}{Z} 
\newcommand{\activlayer}{A} 
\newcommand{\logit}{\lcomb^{[L]}} 
\newcommand{\tweaklogit}{Z^{[L]'}} 
\newcommand{\targetlogit}{Z^{[L]}_t} 
\newcommand{\twtargetlogit}{Z^{[L]^\prime}_t} 
\newcommand{\sensample}{\sample_s} 
\newcommand{\rndsample}{\sample_0} 
\newcommand{\origmodel}{F(\sensample, \weights)} 
\newcommand{\fdelta}{F(\sensample, \weights + \delta)} 
\newcommand{\optss}{\sample^{\ast}} 
\newcommand{\optsens}{\beta^{\ast}} 
\newcommand{\minsens}{\bar{\beta}} 
\newcommand{\minlogit}{\bar{\varepsilon}} 
\newcommand{\neuron}{j} 
\newcommand{\relu}{\textsc{ReLU}}
\newcommand{\reluplex}{Reluplex} 
\newcommand{\dbias}{\mathit{d}\mbox - \mathit{bias}}

\newcommand{\freevar}[1]{\hat{#1}}

\title{%
  Sensitive Samples Revisited: Detecting Neural \\
  Network Attacks Using Constraint Solvers%
  \thanks{Partially supported by the US National Science Foundation under grant \# SaTC-1929300.}}

\author{Amel Nestor Docena \quad Thomas Wahl \quad Trevor Pearce \quad Yunsi Fei
  \institute{Khoury College of Computer Sciences, Northeastern University, Boston, USA}
  \email{$\{$docena.a|t.wahl|pearce.tr|y.fei$\}$@northeastern.edu}}

\def\titlerunning{Detecting Neural Network Attacks Using Constraint Solvers}
\def\authorrunning{A.\ Docena, Th.\ Wahl, T.\ Pearce, Y.\ Fei}

\begin{document}

\maketitle

\begin{abstract}
  Neural Networks are used today in numerous security- and safety-relevant
  domains and are, as such, a popular target of attacks that subvert their
  classification capabilities, by manipulating the network
  parameters. Prior work has introduced \emph{sensitive samples}---inputs
  highly sensitive to parameter changes---to detect such manipulations, and
  proposed a gradient ascent-based approach to compute them. In this paper
  we offer an alternative, using \emph{symbolic constraint solvers}. We
  model the network and a formal specification of a sensitive sample in the
  language of the solver and ask for a solution. This approach supports a
  rich class of queries, corresponding, for instance, to the presence of
  certain types of attacks. Unlike earlier techniques, our approach does
  not depend on convex search domains, or on the suitability of a starting point for the
  search. We address the performance limitations of constraint
  solvers by partitioning the search space for the solver, and exploring
  the partitions according to a balanced schedule that still retains
  completeness of the search. We demonstrate the impact of the use of
  solvers in terms of functionality and search efficiency, using a case
  study for the detection of \emph{Trojan attacks} on Neural Networks.
\end{abstract}

\section{Introduction}

Given recent advances in the field of Deep Learning, Neural Networks (DNN)---the
preferred data structure for many learning tasks---are used today in
numerous application areas, including security- and safety-relevant
domains. Their
use by unsuspecting end users increasingly makes them the target of attacks
that manipulate (a small fraction of) the network parameters, attempting to
override the decision-making functionality of the network, at least for
some inputs. Examples include the hijacking of image recognition software
for access control, and the misguidance of autonomous vehicles. Society has
a vital interest in detecting these kind of attacks, in order to mitigate
or prevent them.

Inspired by recent work by He et al.~\cite{DBLP:conf/cvpr/HeZL19}, we
consider in this paper the ``cloud scenario'': the defender is the designer
of the network, with full access to all parameters and the training
data. To facilitate wide-spread use, after training they deploy the network
using some DNN inference cloud service, which can, however, ultimately not
be trusted. They therefore wish to determine inputs, called \emph{sensitive
  samples}~\cite{DBLP:conf/cvpr/HeZL19}, that are sensitive to parameter
manipulations and thus able to distinguish the original, trusted network,
$N$, from a manipulated one, $N'$.

In the aforementioned recent work, a sensitive sample $\optss$ is defined as
an input that maximizes the difference between $N$ and $N'$, resulting in
an optimization problem. Provided the sample space is convex, $\optss$ can be
found using \emph{(projected) gradient ascent} (PGA). PGA is an efficient
technique, but it is also---from a user perspective---somewhat demanding:
in addition to the convexity of the sample space, we must compute the
differential of the objective function, as well as the projection into the
sample space, both of which can be numerically hard problems.

The goal of this paper is to cast the task of finding sensitive samples as
a Boolean \emph{satisfiability modulo real arithmetic} problem, and use an
SMT solver to crack it. Such solvers do not require convex search spaces,
and they are black boxes: finding a solution for the specified
satisfiability (``sat'') problem is left entirely to the solver.

The flexibility of SMT solving does not come for free. Our sat problem
locks as follows: given the network $N$, find a sensitive sample $\sensample$ such that for all
networks $N'$ s.t.\ $N' \not= N$, we have $N(\sensample) \not= N'(\sensample)$. This is
in fact a sat problem in a \emph{quantified} logic. We approximate it by
restricting the domain of adversarial networks $N'$ to come from some type
of attack commonly applied to $N$, such as a \emph{Trojan
  attack}~\cite{DBLP:conf/isqed/0001M0ZJXS20}. We build a representative
Trojan-attack model $N'$ and obtain a sat problem over quantifier-free
Boolean logic modulo real arithmetic. Input $\sensample$ has qualities resembling
a \emph{test case} for detecting the attack: if positive, the attack is
present; if negative, we cannot fully guarantee the integrity of the cloud
model $N'$.

The second challenge with using SMT solvers is that, given their symbolic
nature, they cannot compete in efficiency with ``concrete''
evaluation-based solvers like PGA-based search engines. The bottleneck in
the SMT solving process is the presence of \relu\ (``rectified linear unit''; see \sectionref{Neural Networks}) activation functions,
which introduce non-linear, non-differentiable arithmetic into the
mathematical model of the neural network (see
also~\cite{DBLP:conf/cav/KatzBDJK17,B21}). We therefore propose in this
paper a (generic) \emph{greedy} algorithm to improve the performance of
sat-solving formulas over many \relu\ instances. Our technique factors the
\relu\ functions out of the formula (reducing its complexity
substantially), and examines the many \emph{cases} that the combination of
\relu\ functions present in a schedule that is determined \emph{a priori} using a
very fast profiling step.

To evaluate our technique, we consider \emph{Trojan attacks}, which turn
the trusted model $N$ into the manipulated cloud model
$N'$~\cite{DBLP:conf/isqed/0001M0ZJXS20}. We demonstrate that (i) our technique
can determine sensitive samples fairly efficiently if used in conjunction
with the greedy algorithm mentioned above, that (ii) these samples
effectively label Trojanned models as such, thus detecting the attack, and
that (iii) \emph{benign} models $N'$ are \emph{not} flagged by our
technique, which would constitute a false positive. A benign model $N'$ is
one that suffers only harmless inference deviations from $N$, not to be
blamed on an attack, such as due to differences in floating-point
arithmetic implementations.

\section{Defender Model and Problem Definition}
\label{section: Defender Model and Problem Definition}

In this work we assume there is a party called \emph{defender} that has
access to a trusted trained machine learning model $N$. The defender seeks to deploy
$N$ to some publicly available DNN query service, which we refer to here simply as \emph{the cloud}, to be
accessed by \emph{end users} via a narrow DNN query API. The cloud provider
has full access to the deployed model (``white-box''); the end user submits
inputs to the service and retrieves a classification result in the form of
probabilities for each class.

Further, there is a party known as \emph{attacker} (often identical to the
cloud provider) set to manipulating the numeric model parameters, i.e., the
weights and biases, resulting in a new model $N'$. The precise goals for
such manipulation vary and include altering the network's functionality,
for some inputs, e.g.,\ using a Trojan
attack~\cite{DBLP:conf/ndss/LiuMALZW018}. As in prior
work~\cite{DBLP:conf/cvpr/HeZL19}, we assume that the attacker does not
manipulate $N$'s hyper-parameters, e.g.,\ by adding extra layers, or adding
neurons to a layer.

We note that, once the defender has deployed the model $N$, they can access
it only via the same narrow interface that is available to standard end
users (``black-box''). That is, the details of $N'$ are unknown to them.

\paragraph{Problem definition.} We address in this paper the following
\emph{idealized} problem for the defender. Given the network $N$ and a type
$T$ of network attacks (such as Trojan attacks), determine a
\emph{detection threshold} $\minsens$ and an input $\sensample$ called
\emph{sensitive sample}~\cite{DBLP:conf/cvpr/HeZL19}, such that the following holds for any
potential cloud model $N'$: if $N$ and $N'$ disagree in their response to
input $\sensample$ by at most $\minsens$, then $N'$ is not the result of an attack
of type $T$ against $N$. In addition, we typically want sample $\sensample$ to be
``similar'' to the inputs given in the training set, so that the use of the
sample by the defender to probe $N'$ does not trigger a ``spy alarm'' by
the cloud provider, which could lead to non-uniform treatment of the
defender, compared to other end users.

The idea is: if input $\sensample$ gives rise to a difference of more than
$\minsens$, networks $N$ and $N'$ disagree non-trivially, which must be
reported, using witness $\sensample$. (For this to make sense, we cannot simply
choose $\minsens=0$; see \sectionref{Sensitive Sample Queries}.)
Otherwise, we consider the cloud model uncompromised, as far as attacks of
type $T$.

The above problem description is idealistic since it contains an implicit
universal quantification over all models $N'$ compromised via a type-$T$
attack. This results in a formula in the expensive (although in principle
decidable) SMT theory of quantified non-linear real arithmetic (NRA). In this
paper we approximate this problem, by determining an input $\sensample$
that is a sensitive sample for a \emph{typical instantiation} $N'$ of a
type $T$-attacked network. This results in a more manageable formula in quantifier-free real arithmetic (QF-NRA). We then check the effectiveness of the sample
thus obtained against other network instances. In general, however, we
cannot guarantee the integrity of the cloud model in the ``otherwise'' case
in the previous paragraph.

\section{Symbolic Specifications of NNs and Sensitive Samples}
\label{section: Network Models and Sample Specifications in SMT}

Our method of choice to tackle the problem defined in the previous section
is via logical constraint solvers. This requires formalizing the neural network (NN), the attack type, and the notion of sensitivity in the language of the solver.
In the Technical Appendix, we give a background of NNs, plus the inner workings of a Trojan attack.

\subsection{Fully-Connected Neural Networks}
Consider an $L$-layer fully-connected NN, (see
Technical Appendix, \sectionref{Neural Networks}). We formulate the linear function
and subsequent non-linear activation function in each hidden layer $l<L$ as
\begin{description}
\item linear combination: $\lcomb^{[l]} = \weights^{[l]\intercal}\activlayer^{[l-1]}$

\item hidden layer activation: $\activlayer^{[l]} = f(\lcomb^{[l]})$,
\end{description}
where $\weights^{\left[l\right]}$, $\lcomb^{\left[l\right]}$,
$\activlayer^{\left[l\right]}$ represent the parameters (weights and bias),
the linear (pre-activation) output vector, and the activated output vector,
respectively. To encode the linear-combi\-nation intermediate result
$\lcomb$ in the SMT-Lib language~\cite{BarFT-SMTLIB}, we declare it to be an uninterpreted
real-valued constant, and then constrain it to be equal to a linear
expression over the components of weights $w_i$ and previous-layer activation $a_i$\,:
\begin{verbatim}
 (declare-const z_h Real)
 (assert (= z_h (+ (* w_1 a_1) (* w_2 a_2)...(* w_n a_n) bias)))
\end{verbatim}

For the activation in hidden layers, \relu\ is a typical choice. We define
it in the SMT-Lib language relationally, as a function $\func {\mathtt{relu}}
{\RR \times \RR} {\BB}$:
\begin{verbatim}
(define-fun relu ((z Real) (a Real)) Bool
  (= a (ite (<= z 0.0) 0.0 z))) ; intuitively, a = relu(z)
\end{verbatim}

For the purpose of adding queries to the encoded neural network, e.g.,\ to
retrieve a sensitive sample, we define the output of the network to be the
logit computed by the pre-activation output function $F$, rather than the result of the output
activation function $\sigma$, which can give rise to complex symbolic
encodings. This is feasible because conditions over the final
activated value $\sigma(F(\sample, \weights))$ can be translated
``backwards'' to conditions over the logit vector, $\logit$. For instance, if
$\sigma = \mbox{sigmoid}$ and we wish to constrain our sample to be
classified as label $t=1$ with probability at least $80\%$, we translate
the condition $\sigma(\logit) \atl 0.8$ to the constraint $\logit \atl
\sigma^{-1}(0.8) = 1.386$ ($\sigma^{-1}$ is called the \emph{logit
  function}).

\subsection{Sensitive Sample Queries}
\label{section: Sensitive Sample Queries}

The \emph{sensitivity} of a sample $\sample$, notated $\beta$, is
measured by the deviation of its prediction when run on a tweaked model
from its original prediction. Selecting some parameters of interest $\poi
\in \weights$ for study, Lee et al.\ define sensitivity as $\beta = \Vert \sigma(\sample, \poi) - \sigma(\sample, \poi + \delta) \Vert$, for some suitable norm $\Vert\cdot\Vert$. A sensitive sample then is an input
$\optss$ that maximizes this sensitivity assuming that the $\poi$ have been
tweaked by some $\delta$: $\optss = \argmax_I \Vert\sigma(I, \poi) -
\sigma(I, \poi + \delta)\Vert$ \cite{DBLP:conf/cvpr/HeZL19}. For this
$\optss$, its corresponding sensitivity $\optsens$ is optimal.

In this paper, instead of solving an optimization problem, we determine
input samples that give rise to a suitable ``target
sensitivity''. Requiring this sensitivity to be positive is not enough:
differences in the compiler, the computation environment, the available
hardware and other unknowns (which impact the precise semantics of
floating-point arithmetic \cite{DBLP:conf/europar/GuWBL15}) will typically
cause some deviations in the output between the client's platform and the
cloud. In the absence of an attack, these deviations would show up as false
positives. We therefore model floating-point vagaries using an
application-dependent \emph{detection threshold} $\minsens$ beyond which
any observed sensitivity is blamed on the presence of an attack, while
sensitivities below it are assumed to be harmless. We show later in our
experiments that such deviations tend to be far
smaller than differences due to an attack, making the two quite easily
separable via a suitable detection threshold. Thus, an input $\sensample$ witnesses the presence of an attack iff its sensitivity $\beta_s \atl \minsens$.

In contrast to PGA where the sensitivity of a sample is determined when a SS is retrieved, (if the model fails to converge, however, then no optimal SS is retrieved); in our approach, we set a threshold for the sensitivity first and determine whether we can find a satisfying assignment for our prescribed SS. In the following subsection we spell out an attack symbolically to determine a desired $\sensample$, displaying flexible specifications, leading to a case where we detect a Trojan attack.

\subsubsection{Symbolically Encoding Sensitive Sample Conditions}

Consider an attack on weight parameters whose goal is to manipulate the original value of some target logit $\targetlogit \in \logit$ to a desired value $\twtargetlogit$ and, consequently, to belie the prediction. In detecting this attack, we select $\poi \in \weights$ on which we assume weight changes, $\delta$. These $\poi$ are referred to as the \emph{parameters of interest} (POI) \cite{DBLP:conf/cvpr/HeZL19}, selected based on knowledge about an attack, which are representative of the actual parameters that have been perturbed for detection. We specify the conditions for the sensitive sample as follows:

\begin{equation}
  \label{equation: Approach2Variant}
  \begin{array}{c}
    \sensample = \rndsample + \freevar{\gamma} \ \land \ \sensample \in S \ \land \ \sensample \models C \ \land \ \logit=\origmodel \land \ \\
    \tweaklogit = \fdelta \ \land \ \forall \delta_i: \ \delta_i \in [a,b]_i \ \land \ \left |\twtargetlogit - \targetlogit \right| \atl \minlogit \ .
  \end{array}
\end{equation}

The sensitive sample $\sensample$ takes the form $\sensample = \rndsample + \freevar{\gamma}$, where $\rndsample$ is a training sample modified by some suitable transform variables $\freevar{\gamma}$. We distinguish the variable we seek for satisfiability from the rest of the parameters that the defender defines to bound the search with a hat; in the case of \equationref{Approach2Variant}, these are the $\freevar{\gamma}$. While, the variables that are given or defined by the defender are: the training sample, $\rndsample$; the sample space $S$ and additional constraints $C$; the DNN architecture, $F(\cdot)$; the trained weights, $W$; the weight change vector $\delta$ applied to the POI and its assumed range of values; and the logit detection threshold $\minlogit$, which sets the sensitivity of the sample that satisfies the detection threshold. We expound on these conditions further. 
	\par By constraining $\freevar{\gamma}$ to within a small radius around 0, we force the SS to be close to the training sample $X_0$; thereby, appear ``natural'', not artificial, to an input analyzer that may be used by the service provider to detect whether their inference is being monitored or tested. Prior work has used similar
mechanisms to enforce similarity of the sensitive sample to the training
data~\cite{DBLP:conf/cvpr/HeZL19}. Moreover, the sample space $S$, which the SS is an element of, can be convex or non-convex, as can be the additional
constraints~$C$. Such constraints might state that the sample should be
classified into a particular label by the network. Further, when $\sensample$ is
forward-propagated through the network, the output, expressed as a formula over the logit vector $\logit$, must of course agree
with the network formula $\origmodel$, a function of $\sensample$ and $\weights$.

\par For our assumed tweaked model---$\fdelta$, a function of $\sensample$, the $\weights$ and corresponding weight-change vector $\delta$---we are assuming a reasonable range for the weight perturbations $\delta_i$ (components of $\delta$) applied to the POI. The choice of POI and range of weight-change is attack-dependent. The actual weight deltas are not known to us, but we know the range of the trained weights. We can therefore get a sense of a plausible range of the deltas based on the nature of the attack (later we present suitable choices for the case of a Trojan
attack). Note that we are expressing sensitivity in terms of the logits. We refer to $\minlogit$ as the \emph{satisfying logit threshold}, whose
equivalent sensitivity (measured in terms of the output probability, by
applying $\sigma$ to the logits) satisfies the detection threshold
$\minsens$. In \sectionref{Evaluation}, we present an example to
configure this metric that models sensitivity given a detection threshold.
\par So in a nutshell, we wish to solve for $\freevar{\gamma}$ such that the sensitive
sample $\sensample$ captures a bandwidth $\minlogit$ when the POI have
been tweaked by $\delta$. In the next specification where we take on a Trojan attack, the parameters assumed as placeholders take shape.

\subsubsection{Detecting a Trojan Attack}
\label{section: Detect trojans}

After a Trojan attack, \cite{DBLP:conf/ndss/LiuMALZW018}
observed that some weights from the target layer through the output layer
will be inflated, causing a jump in the output towards the target masquerade $t$; while, the rest will be re-adjusted to compensate for the inflation, making the Trojanned model to behave like the original model when the trigger is absent. To devise sensitivity conditions to detect this attack, we translate these observations as a special configuration of \equationref{Approach2Variant}, as explained in the following.

We detect whether our model has been Trojanned towards some label $t$,
which we select for scrutiny. As POI for this attack, we choose the weights connected to the output layer (a~similar strategy was employed
in~\cite{DBLP:conf/cvpr/HeZL19}); the weights from the target layer (which
only the attacker knows) all the way up to just before the output layer are
assumed unchanged. Among the POI, we let $\poi^{L}_e$ be the weights expected to be inflated; while, the rest of the weights potentially deflated as $\poi^{L}_d$. That is: we
assume positive weight deltas $\delta_e > 0$ applied to the former weights,
and non-positive weight deltas $\delta_d \leq 0$ applied to the latter. The
corresponding activated neurons connected by these weights are $\activlayer^{[L-1]}_e$
and $\activlayer^{[L-1]}_d$, respectively. We formulate the original logit
for label $t$ as $\targetlogit = \poi_e^{[L] \intercal}
\activlayer^{[L-1]}_e + \poi_d^{[L] \intercal}
\activlayer^{[L-1]}_d$. The perturbed logit, with the corresponding weight
deltas, is given by
\[
  \twtargetlogit = (\poi_e^{[L]} + \delta_e)^{\intercal} \activlayer^{[L-1]}_e + (\poi_d^{[L]} + \delta_d)^{\intercal} \activlayer^{[L-1]}_d \ .
\]
Since a Trojan attack raises the prediction towards the target masquerade $t$ in
the presence of a trigger, we set the difference between the perturbed
logit and the original logit to be positive and non-trivial,
$\minlogit>0$. Upon subtraction, we get $\twtargetlogit - \targetlogit =
\delta_e^{\intercal} \activlayer^{[L-1]}_e + \delta_d^{\intercal}
\activlayer^{[L-1]}_d = \delta^{\intercal} \activlayer^{[L-1]} \geq
\minlogit$. This suggests that we model our sensitive sample to have a
non-trivial net sensitivity on the possible weight perturbations that can occur among the POI in a Trojan attack. With this as sensitivity condition, the sensitive-sample specification from \equationref{Approach2Variant} becomes:
\begin{equation}
  \label{equation: Trojan detection}
  \begin{array} c
    \sensample = \rndsample + \freevar{\gamma} \ \land \ \sensample \in S \ \land \ \sensample \models C \ \land \ \logit=\origmodel \ \land \\ 
    \tweaklogit = \fdelta \ 
    \land \ \delta_d \atm 0 < \delta_e \ \land \ \delta^{\intercal} \activlayer^{[L-1]} \atl \minlogit \ .
  \end{array}
\end{equation}

In \sectionref{Evaluation}, we determine suitable values for these parameters that bound $\freevar{\gamma}$ by detecting a real-world Trojan attack. But before that, we devise an algorithm to tackle scalability of this approach in the next section.

\section{Improving Scalability using \relu\ Profiling}
\label{section: Improving Scalability using relu Profiling}

For non-trivial networks, the SMT models designed in \sectionref{Network
  Models and Sample Specifications in SMT} represent hard decision
problems. In this section we get to the bottom of the complexity, and
design an algorithm to improve the scalability.

\subsection{Root-Cause Analysis: Scalability Bottleneck}

We analyze the root cause of the scalability problems. As also observed in
other work~\cite{DBLP:conf/cav/KatzBDJK17}, the main culprit is the
``branches'' that each application of a \relu\ activation function
represents: they cause the network model to be a piece-wise linear, rather
than linear, mathematical function of the inputs. We
can (vastly) underapproximate the SMT model for the sample query, by
replacing each \relu\ activation function by either the identity or the
constant-zero function---we say the \relu\ node is \emph{fixed as identity}
or \emph{fixed as zero}--- and simultaneously forcing the inputs to the
function to be respectively non-negative or negative:
\begin{verbatim}
  ; id: R x R -> Boolean
  (define-fun id ((z Real) (a Real)) Bool
    (and (>= z 0) (= a z)))                 ; z >= 0 and a = z

  ; zero: R x R -> Boolean
  (define-fun zero ((z Real) (a Real)) Bool
    (and (< z 0) (= a 0))                   ;  z < 0 and a = 0
\end{verbatim}
Performing this replacement on one \relu\ node roughly cuts the solution
space for the solver to explore in half. We now design an algorithm for more efficient
sensitive-sample search exploiting these insights, and demonstrate its
impact in \sectionref{Evaluation}.

\subsection{Greedy ReLU Branch Exploration using Decision Profiling}

Solving a query for a sample input intuitively requires exhaustively
exploring all combinations of branch decisions that the \relu\ nodes might
make---\emph{\relu\ combinations} for short---for a given input. The number
of such combinations is, of course, exponential in the number of
\relu\ nodes, resulting in poor scalability. A key idea, however, is that
we are free to choose the \emph{order} in which the combinations are
examined. An optimistic approach might try first \relu\ combinations that
are deemed ``more likely'' to permit a satisfying assignment, \emph{easier
  to solve} for short. Completeness of this approach can be guaranteed by
exploring \emph{harder to solve} combinations later, rather than discarding
them.

\vspace{3pt}

\noindent
\centerTwoOut{%
  \begin{minipage}[c]{0.55\textwidth}
    \quad\quad But what makes one \relu\ combination easier than others? We
    borrow here the idea of \emph{branch prediction} done by
    runtime-optimizing compilers: As the owner of the network model and the
    training data used to obtain it, we can perform an inexpensive
    \emph{decision profiling} step, which determines, for each \relu\ node,
    how often it acts as the identity function, and how often as the zero
    function, measured over the training data. We call the larger of these
    two
  \end{minipage}}{%
  \begin{minipage}[c]{0.4\textwidth}
    \includegraphics[scale=0.4]{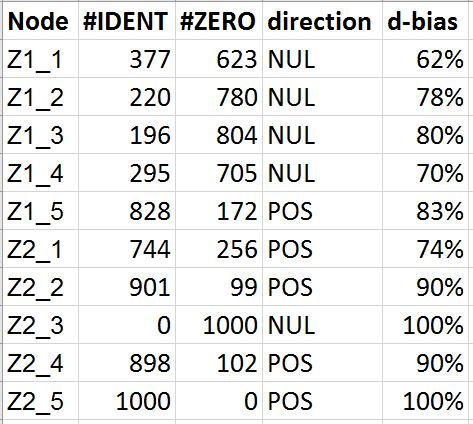}
  \end{minipage}}

\vspace{3pt}

\noindent
numbers, as a percentage of the training data size, the \emph{decision
  bias}, \emph{d-bias} for short, of a \relu\ node: a large d-bias towards
``identity'' suggests the node acts more often as the identity function
than as the zero function. The table above illustrates, for a toy network
of two hidden layers with 5 neurons each and a dataset of 1000 inputs, the
number if times a \relu\ node acted as identity or as zero, the direction
of the bias, and the d-bias value (formally defined
in~\equationref[]{d-bias} below).

Since the sensitive samples are designed to be (slight) perturbations of
the training data, we expect it to be beneficial to assume that the
\relu\ nodes exhibit a branching behavior similar to that over the training
data. We say a \relu\ node has been \emph{fixed} if it has been fixed
according to its d-bias (ties resolved in some arbitrary but deterministic
way). To \emph{unfix} a (fixed) \relu\ node means to reinstate the
\relu\ function, in place of the identity or zero function that it had been
replaced with.

Equipped with these definitions, we deem a \relu\ combination easier to
solve than another if the \relu\ nodes that have been fixed in the former
form a superset of those fixed in the latter. Additionally, we sort the
\relu\ nodes according to their d-bias and use this ordering to make
\relu\ combinations easier to solve. The motivation is that fixing a
\relu\ node with a large d-bias is more likely to preserve many satisfying
solutions than fixing a \relu\ node with a small d-bias (near $50\%$).

To formalize these ideas, let $(\range[]{\sample_1}{\sample_k})$ be an input to the
network. Consider a neuron $\neuron$, and let $\lcomb_\neuron$ be the function that computes the pre-activation value of the neuron, i.e.,\ the input to the
\relu\ function at $\neuron$. (The output computed at $\neuron$ is therefore
$\activlayer_\neuron = \relu(\lcomb_\neuron(\range[]{\sample_1}{\sample_k}))$.) We say $\neuron$ \emph{acts as identity on $(\range[]{\sample_1}{\sample_k})$} if $\lcomb_\neuron(\range[]{\sample_1}{\sample_k}) \atl 0$; otherwise $\neuron$
\emph{acts as zero} on this input. For a set $\data$ of network inputs (such as
a training set), the \emph{d-bias} of neuron $\neuron$ (a number in $[0.5,1]$) is
defined~as
\begin{equation}
  \label{equation: d-bias}
  \dbias(\neuron) =
  \max
  \left\{
  \begin{array}{l}
    |\{(\range[]{\sample_1}{\sample_k}) \in \data: \lcomb_\neuron(\range[]{\sample_1}{\sample_k}) \atl 0\}| , \\[3pt]
    |\{(\range[]{\sample_1}{\sample_k}) \in \data: \lcomb_\neuron(\range[]{\sample_1}{\sample_k}) <    0\}|
  \end{array}
  \right\} \wbox{$/$} |\data| \ .
\end{equation}

\algorithmref{greedy} implements the \relu-aware search strategy we
sketched above. It~takes as input a NN model $\origmodel$ along with a training data
set $\data$, and a formula $\phi$ over the model inputs $\range[]{\sample_1}{\sample_k}$.
Formula $\phi$ typically encodes some kind of condition for an input that
we are trying to find, e.g.,\ the condition of $\range[]{\sample_1}{\sample_k}$ being a
sensitive sample for $\origmodel$. The algorithm begins by computing the d-biases of
all \relu\ nodes over the training set. It then fixes each \relu\ node
according to its d-bias, i.e.,\ it replaces, in $\phi$, each
\relu\ activation function by the identity or the zero function, depending
on the direction of the bias.

If the modified formula $\phi$, which represents an underapproximation of
the original $\phi$, permits a solution, this solution is genuine, and we
return it. Otherwise, we have to weaken the formula, by unfixing one of the
fixed \relu\ nodes. Here we unfix nodes with \emph{small} d-bias first: a
small bias means the predictive power of the decision profiling is weak, so
unfixing this node may re-enable many solutions. This order heuristic is
implemented via the sorting step in \lineref{sort}; ties are resolved
arbitrarily.
\begin{algorithm}[htbp]
  \begin{algorithmic}[1]
    \REQUIRE{network model $\origmodel$, training data $\data$, formula $\phi$}
    \ENSURE {a model input $\range[]{\sample_1}{\sample_k}$ satisfying $\phi$, if one exists; otherwise ``unsat''}
    \STMT compute the d-biases of all \relu\ nodes in $\origmodel$ w.r.t.\ data set $\data$
    \STMT in $\phi$, fix each \relu\ node according to its d-bias
    \STMT sort the \relu\ nodes in $\origmodel$ in order of \emphasize{\em in}creasing d-bias: $\range[]{\activlayer_1}{\activlayer_l}$ \label{line: sort}
    \FOR {\code{$\neuron$ := 1 to $l$}} \label{line: for}
      \IF{there exists $\range[]{\sample_1}{\sample_k}$: $(\range[]{\sample_1}{\sample_k}) \models \phi$}
        \STMT \plreturn\ ``solution: $\range[]{\sample_1}{\sample_k}$''
      \ENDIF
      \STMT unfix \relu\ node $\activlayer_\neuron$ \label{line: unsat: underapproximation}
    \ENDFOR
    \STMT \plreturn\ ``unsat''
  \end{algorithmic}
  \caption{Greedy \relu\ branch exploration using decision profiling}
  \label{algorithm: greedy}
\end{algorithm}

We emphasize that we have merely used heuristics to determine the
\emph{order} in which different \relu\ combinations are
searched. Theoretical completeness of the algorithm is not affected, since
all combinations are, in the worst case, examined. If we were to pass
$\phi$ directly to the SMT solver, we would leave it to the solver to
examine these combinations, oblivious to the information offered by the
profiling~data.

\section{Evaluation}
\label{section: Evaluation}
\overfullrule = 0pt
We conducted experiments to retrieve samples sensitive against a Trojan attack, as motivated in \sectionref{Detect trojans}, and
checked their effectiveness in detecting the attack. We then assessed the
scalability of the technique to larger networks, both without and with
\algorithmref{greedy}. We used Microsoft's Z3 as SMT solver. The
experiments were run on an Intel Core i7-10750H CPU at 2.60GHz and 16GB
of~RAM.

\subsection{Victim Network}
Our benchmarks for detecting a Trojan attack come from the \emph{Belgium
  Traffic Signs dataset}
\cite[\url{https://btsd.ethz.ch/shareddata}]{DBLP:conf/ijcnn/MathiasTBG13}. We
re-sized the images to 14x14 pixels and turned them grayscale. We trained a fully-connected NN of dimensions 30x20x10x1 to identify whether a traffic
sign indicates a speed limit (label 0) or STOP (label 1). For this
mini-image classification task, this NN---despite being small---has
100\% validation accuracy, precision, and recall.

\subsection{Configuring the parameter space}
\label{section: parameter space}
We partition the sensitive-sample parameter space into \emph{attack parameters} and \emph{sample parameters}. Our configuration was defined independently of the attack simulation.

\paragraph{Attack parameters.} They model the Trojan attack and include the
following:
\begin{description}
\item[\normalfont Assumed target masquerade, $t$:] STOP-sign label.

\item[\normalfont POI, $\poi \in \weights$:] weights attached to the output layer.

\item[\normalfont Weight deltas, $\delta$:] We assume the top weights, in terms of value, to be inflated and few, since the attack is supposedly stealthy; the rest are assumed potentially deflated, (i.e., non-positive delta). For this experiment where the trained weights are within $[0, 1]$, we assumed 30\% are inflated by $[0.05,0.25]$ units; (we are not setting a larger sub-range since we are modelling an attack that is stealthy).

\end{description}

\paragraph{Sample parameters.} They include the sample space and the
detection threshold. We further added as constraint the predicted label of the sample on the original model.

\begin{description}
\item[\normalfont Sample space:] In order to make the sensitive sample
  appear like a regular input, we randomly picked a speed limit sign $\rndsample$
  from the test data (see \figureref{SSResult} (left)) and added transform
  variables, $\freevar{\gamma}$, over the entire region of the pixel dimensions. The SS is the
  pointwise sum of the pixel values of $\rndsample$ and the assignment to the
  transform variables.

\item[\normalfont Predicted output label:] We explicitly required that the
  originally predicted output label remain as a speed limit sign even after
  the SS transformation.

\item[\normalfont Detection threshold, $\minsens$, and the satisfying logit threshold, $\minlogit$:] We first stipulated a detection threshold that is significant to warrant weight perturbations: we set $\minsens = 0.01$; that is, a sensitivity of 1 percentage point in probability output or more is attributed to an attack in model parameters. Based on this requirement, we determined $\minlogit$ by computing the initial logit and then setting a satisfying logit threshold. By forward propagation on the original network, we computed the logit of the random sample $\rndsample$ to be $Z^{[4]}=-4.8510$. In this experiment, we set $\minlogit = 1$, which models a tweaked logit $Z^{[4]\prime} \atl -3.8510$. The corresponding modeled sensitivity under this setting is $\beta \atl 0.0131$, which satisfies the detection threshold.
\end{description}

\subsection{Effectiveness of SS}
After solving for the transform values, we tested the obtained sensitive
sample in detecting the Trojanned model. We deem the sample effective if
the observed sensitivity is at least the detection threshold. If it is below, this would be a false negative. Furthermore, to assess the
possibility of false positives, we subjected the original model to minor
perturbations as they may occur ``innocently'' on the cloud. Concretely, we compared the prediction of the sensitive samples on a version where model parameters are originally stored in \emph{float16} precision to one where the parameters are stored in \emph{float32}. If the sensitivity to such innocent perturbations (SIP) remains below the detection threshold, then the sample does not represent a false positive.

\subsection{Results}

\paragraph{Naive approach:} This approach passes \equationref{Trojan
  detection} to the SMT solver (w/o \algorithmref{greedy}). After
solving for the transform variables (which took about one minute), we
obtained the sample shown in \figureref{SSResult} (right): it has a
sensitivity of $\beta = 0.0744$ (summarized in the first row of
\tableref{SearchWOReluProfile} in the Appendix). This sensitivity means that the prediction
of the Trojanned model is 7.44 percentage points away from the original
prediction. Moreover, given $\mathit{SIP} = \mbox{7.41E-06}$, the sample
does not exhibit a false positive.

  \begin{figure}[htbp]
    \centering
    \includegraphics[scale=0.40]{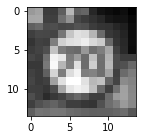}
    \quad \quad
    \includegraphics[scale=0.40]{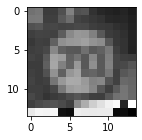}
    \caption{Randomly taken test sample (left) and transformed, \emph{sensitive} sample (right)}
    \label{figure: SSResult}
  \end{figure}

\paragraph{Scaling of \equationref[]{Trojan detection} to larger networks:} \sectionref{Scalability of naive sample search via SMT} of the Technical Appendix shows results of attempting to scale the naive approach to larger networks. \sectionref{Raw size data for encodings of various networks} shows raw size data for various SMT encodings.

\paragraph{Employing \algorithmref{greedy}:} We allotted 20/60/120sec as
the increasing time per iteration of the \plkeyword{for} loop in
\lineref{for}, before the algorithm moves on to the next
\relu\ combination. This reflects the greedy character of the algorithm: we
want to first try many instantiations of \relu\ nodes as ``identity'' or
``zero'', believing that one of them will give us a solution. If this
approach fails, we increase the time allotment per iteration. This serves
the goal of delaying advancing to later iterations, with fewer fixed
\relu\ nodes, as much as possible, which eventually degenerates the
algorithm to one that simply has the solver explore the many
\relu\ combinations. In \tableref{SearchWReLUProfile}, the iteration number
where a satisfying assignment was finally found is given in column
``\textbf{iter.~\#}''. Small numbers here indicate success of the
philosophy purported by \algorithmref{greedy}. Column
``\emphasize{runtime}'' shows the total time it took to find a satisfying
assignment, i.e.,\ the sum across all iterations, for a respective iteration
timeout of 20/60/120sec. If a solution is found for an iteration timeout of 20sec, one would normally stop the algorithm. In order to evaluate
scalability, we include results for larger time allotments. But despite the extension, the solver wounded up with the same result as with the initial allocation.

\begin{table}[htbp]
  \centering
  \begin{tabular}{| c | r | c | c | c | r |}
    \hline
    \textbf{Network size} & 
    \textbf{iter.~\#} & 
    \textbf{sensitivity} & 
    \textbf{SIP} &
    \mc 1 {c|} {\textbf{total runtime (s)}} \\
    \hline
    30x20x10x1 & 6 & 0.1234 & 1.08E-05 & 111/312/\ 612 \\
    \hline
    40x20x20x1 & 4 & 0.8342 & 2.99E-05 & \ 65/185/\ 365 \\
    \hline
    50x30x10x1 & 14 & 0.3520 & 1.33E-05 & 264/784/1564 \\
    \hline
    50x30x20x1 & 12 & 0.0249 & 4.00E-05 & 230/670/1330 \\
    \hline
    60x40x20x1 & 2 & 0.9859 & 5.04E-06 & \ 36/\ 75/\ 135 \\
    \hline
    90x60x30x1 & --- & --- & --- & (out of resources) \\
    \hline
  \end{tabular}\hspace*{-25pt}
  \caption{Search with ReLU profiling algorithm}
  \label{table: SearchWReLUProfile}
\end{table}

\paragraph{Comparison naive approach/\algorithmref{greedy}:} Consider the
case of the 30x20x10x1 NN model. \algorithmref{greedy} found a satisfying
assignment in iteration \#6. This means that the activation functions of
five neurons---those with the smallest d-bias---were freed from their fixed
instantiation to the identity or zero functions, and reverted to exact
\relu\ semantics. This combination yielded a sensitivity of 12.34
percentage points, a higher sensitivity than the solution found by the naive approach. The time reported for
\algorithmref{greedy}'s 20s/iter.\ run (111s) is larger though. This is not
surprising, since \algorithmref{greedy} ``wastes'' five SMT runs. In fact,
the motivation for using this algorithm is not to find solutions in
``easy'' cases faster. It is, instead, to increase scalability to larger
models. Indeed, \algorithmref{greedy} was able to solve all models except
the 90x60x30x1 case, while the naive approach timed out for most.

\par Presently, the method is not able to handle deeper and larger networks, such as state-of-the-art convolutional neural networks. Nevertheless, the DNN models that we presented are valid networks; we showed that SMT solvers can be used for such NN queries, e.g., sensitive-sample generation, when appropriately formalized. To deal with the complexity of DNNs, research has been conducted into dedicated theories for NN queries, e.g., \reluplex~\cite{DBLP:conf/cav/KatzBDJK17}. Using a dedicated solver may potentially address the scalability issue further: we first use \algorithmref{greedy} to eliminate some ReLU nodes, (e.g. the ones with high bias), while others are left in. For those left in, \reluplex~can be applied. The investigation of this possibility can be picked up in future work.

\section{Related Work}
This work was inspired by the \emph{sensitive-sample fingerprinting}
technique proposed by He et al.~\cite{DBLP:conf/cvpr/HeZL19}, which uses
classic machine learning techniques based on (projected) gradient ascent to
determine sensitive samples. This is very efficient, but requires a
starting point for the search and a \emph{convex} solution
space. Lack of convexity can lead to
sub-optimality or, worse, failure to converge. The goal here was to solve a
similar problem, but bring the flexibility of symbolic constraint solvers
to bear: we can specify any search space and sample conditions, as long as
they are definable in the underlying logic. However, definability does not
imply efficient processibility, which is why \sectionref{Improving
  Scalability using relu Profiling} presents an algorithm for improved
satisfiability checking.

Solving an optimization problem, the gradient-based technique returns
samples that maximize the sensitivity. The authors conclude that any output
discrepancy confirms the presence of an attack (``guarantee[s] zero false
positives'' \cite{DBLP:conf/cvpr/HeZL19}). As discussed in
\sectionref{Sensitive Sample Queries}, this is not quite true: due to
platform-dependencies of floating-point computations
\cite{DBLP:conf/europar/GuWBL15}, DNN model inference is \emph{not}
deterministic. We address this problem using an empirical non-zero
sensitivity detection threshold (\sectionref{Defender Model and Problem
  Definition}).

Using symbolic techniques in deep learning is still a relatively young
area; examples
include~\cite{DBLP:conf/tacas/GiacobbeHL20,DBLP:journals/aicom/PulinaT12}. The
\reluplex\ SMT solver~\cite{DBLP:conf/cav/KatzBDJK17} introduces a
theory of real arithmetic extended by the \relu\ function as a primitive
operation. We can view our greedy \algorithmref{greedy} as an alternative
dedicated way of handling \relu\ nodes. A~stand-alone comparison of both
methods is left for future work.

Our method needs to be instantiated for different attack types (to avoid an unrealistic universal quantification over ``all'' adversarial models). We
have focused on Trojan attacks, a survey can be found
in~\cite{DBLP:conf/isqed/0001M0ZJXS20}. Strategies for formalizing other types of DNN attacks are left as future work. ``Fingerprinting'', using inputs
characteristic of model manipulations, is one way of detecting such
attacks; there are others. For instance, we can conclude the model has been
compromised if, for some class, the minimal trigger that causes a
misclassification is substantially smaller than for other
classes and thus is likely supported by a
  Trojan~\cite{DBLP:conf/sp/WangYSLVZZ19}. In that approach, the defender
does not need access to the trusted model or the training data. The
approach is designed specifically for backdoor-style attacks relying on a
trigger. Other methods perform statistical analyses, e.g.,\ determining the
probability distribution of potential
triggers~\cite{DBLP:conf/ijcai/ChenFZK19} or of prediction results under
perturbations~\cite{DBLP:conf/acsac/GaoXW0RN19}. Such analyses may not be
realistic in a cloud environment.

\section{Summary}

We revisited the technique of retrieving sensitive samples in detecting NN attacks. A previous approach solves an optimization problem, using an
efficient projected gradient-based search, in order to find the optimal
sensitivity~\cite{DBLP:conf/cvpr/HeZL19}, which faces, however, various
technical preconditions and is somewhat rigid. Our approach performs the search via a symbolic constraint solver. This
permits a flexible specification of desirable features of the sample. We
argue that a sample need not be optimal to be effective, as long as its sensitivity is above a
threshold that delineates it from innocent perturbations that can occur
upon upload of the NN model to the cloud. This
alternative comes with the price of efficiency, however. To address
scalability, we devised a greedy algorithm that searches through all
possible combinations of the ReLU node behaviors in an ``easiest-first''
order. This algorithm has applications in symbolic NN analysis beyond
sensitive-sample search. Future work includes investigating the performance impact of using dedicated solvers for NN queries, such
as~\reluplex~\cite{DBLP:conf/cav/KatzBDJK17}.

\bibliographystyle{eptcs}
\bibliography{bibliography}

\appendix

\section*{Technical Appendix}
\section{Background}
\subsection{Neural Networks} \label{section: Neural Networks}

A Neural Network (NN) is a parameterized, layered function that maps a
vector $\sample$ of features from some $n$-dimensional input space to an
output $y$, which may be discrete if the task is classification, or real if
regression. Given parameters $\weights$ (a matrix of weights and biases), each layer
consists of a linear function $l_i$ over the layer's inputs and its weight
parameters, and a subsequent non-linear activation function $f_i$. Function
$l_i$ is a simple dot product, translated by the bias, for a fully-connected layer. The activation for inner layers is typically implemented via the
\emph{rectified linear unit}, defined as $\relu(b) = \max\{b,0\}$. The
activated values become the inputs of the next layer. The result of the final linear function, known as \emph{logit},
i.e.,\ computed at the output layer, is passed to a special activation
function $\sigma$. For a binary classification task, the output activation
is typically a sigmoid function; for a multi-label classification, it is a
softmax function. Output $y$ is then a set of probabilities indicating
which among the labels the input $\sample$ most likely belongs to. For a
regression task, the output activation is a linear function, which yields
an output over the real numbers. Thus given a network of $L$ layers, we summarize the function computed by the NN
as
\begin{center} 
  $y \ = \ \sigma(l_L(f_{L-1}( \ldots f_1(l_1(\sample,\weights)) \ldots ))) \ = \ \sigma(F(\sample, \weights)), \ $
\end{center}
where $F$ denotes the pre-activation output, i.e.,\ the logit value computed
for input $\sample$.

\subsection{Trojan Attack on Neural Networks}
A \emph{Trojan} attack on NN perturbs model parameters in order to
cause a misclassification towards a target label on inputs with an embedded trigger; the modified
network is said to be \emph{Trojanned}. But, when the
Trojanned model is presented with inputs without the trigger, the
prediction is unchanged. This makes the attack hard to detect by
unsuspecting users.

\vspace{3pt}

\noindent
\centerTwoOut{%
  \begin{minipage}[c]{0.55\textwidth}
    \quad\quad There are three steps to Trojan a neural network (see figure
    on the right, taken from~\cite{DBLP:conf/ndss/LiuMALZW018}). The
    attacker is assumed to be able to modify the model parameters. First,
    the attacker generates a Trojan trigger, which is a snippet embedded to
    a test input that excites certain neurons so that the prediction is
    skewed towards a target masquerade. This trigger is initialized with a
    mask or shape. \cite{DBLP:conf/ndss/LiuMALZW018} suggests that the
    target layer from which to establish a link with the trigger is
    somewhere near the middle layer, wherein the neurons that are
  \end{minipage}}{%
  \begin{minipage}[c]{0.42\textwidth}
    \includegraphics[scale=0.37]{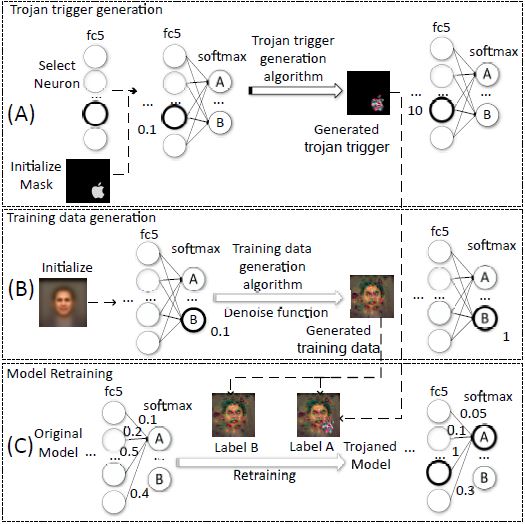}
  \end{minipage}}

\vspace{3pt}

\noindent
most connected are selected.

The authors define weight connection of a neuron as the L1-norm of the
weights connected to the neuron from the previous layer. The values of the
initialized trigger are set by optimizing the activation values of the
select neurons to intended values, supposedly large. This establishes a
strong connection between the trigger and the select neurons, so that in
the presence of the former the latter have strong activations, which would
influence the model prediction.

After generating the trigger, the attacker stamps it to some training
samples for each of the output labels, wherein the label for all these
stamped samples is the target masquerade. If the attacker does not have
access to the training data, they can generate the latter by model
inversion. The batch of samples to be used for retraining the NN would
consist of both unstamped and stamped samples. This last step of retraining
the original NN Trojans the network, such that the weights are re-adjusted
so that the new model predicts the target masquerade whenever the trigger
is present in a test input, but predicts normally when absent, making the
attack stealthy.

\section{Scalability of naive sample search via SMT}
\label{section: Scalability of naive sample search via SMT}
Using the same training data and basic network architecture, we trained larger networks, by expanding the layer sizes. We then simulated a Trojan attack on each of these. In order to assess scalability, we kept all SS parameters as set earlier. (In practice, one might determine individual SS parameters for each network.) Although we were able to transform sensitive samples in some networks,
\tableref{SearchWOReluProfile} shows the poor scaling of the naive approach
to larger networks, especially where the size of the layers close to the
output increases. 

\begin{table}[htbp]
  \centering
  \begin{tabular}{| c | c | c | c |}
    \hline
    \textbf{Network size}  & \textbf{sensitivity} & \textbf{SIP} & \textbf{runtime} \\
    \hline 
    30x20x10x1 & 0.0744 & 7.41E-06 & 71s \\
    \hline
    40x20x20x1 & --- & --- & (out of resources) \\
    \hline
    50x30x10x1 & 0.0595 & 8.85E-07 & 4 \\
    \hline
    50x30x20x1 & --- & --- & (out of resources) \\
    \hline 
    60x40x20x1 & --- & --- & (out of resources) \\
    \hline
    90x60x30x1 & --- & --- & (out of resources) \\
    \hline
  \end{tabular}
  \caption{Sensitivities and running time for sample search without \relu\ profiling}
  \label{table: SearchWOReluProfile}
\end{table}

\section{Raw size data for encodings of various networks}
\label{section: Raw size data for encodings of various networks}

To convey an idea of the raw complexity of the SMT approach,
\tableref{NetworkFormulation} shows various size data of the network models
and the SMT encoding \equationref[]{Trojan detection} of the SS search.
\begin{table}[htbp]
  \small
  \centering
  \begin{tabular}{| c | r | r | r | r |}
    \hline
    \textbf{NN dimensions} & \textbf{\# NN param.} & \textbf{\# \relu\ nodes} & \textbf{SMT: LOC} & \textbf{SMT: file size} \\
    \hline
    30x20x10x1 & 6,751 & 60 & 7,853 & 354kB \\
    \hline
    40x20x20x1 & 9,141 & 80 & 10,363 & 468kB \\
    \hline
    50x30x10x1 & 11,701	& 90 & 12,943 & 592kB \\
    \hline
    50x30x20x1 & 12,021	& 100 & 13,333 & 613kB \\
	\hline
    60x40x20x1 & 15,101 & 120 & 16,503 & 770kB \\
    \hline
    90x60x30x1 & 25,051 & 180 & 26,753 & 1,284kB \\
    \hline
  \end{tabular}
  \caption{Complexity of SMT approach: NN dimensions, number of weights and
    biases, number of ReLU nodes, lines of code of the SMT encoding, and
    file size}
  \label{table: NetworkFormulation}
\end{table}

\end{document}